\documentclass[runningheads]{llncs}
\usepackage[utf8]{inputenc}
\usepackage[]{todonotes}
\usepackage{amssymb,amsmath}
\usepackage[all]{xy}
\usepackage{graphicx}
\usepackage{url}
\usepackage{amsfonts}
\usepackage{mdframed}
\usepackage{url}
\usepackage{hyperref}
\usepackage{algorithm} 
\usepackage{algpseudocode}
\usepackage{bm}
\usepackage{listings}
\usepackage[inline]{enumitem}
\graphicspath{{./img/}{./}}

\newcommand\cA{\ensuremath{\mathcal{A}}\xspace}

\newcommand\cE{\ensuremath{\mathcal{E}}\xspace}

\newcommand\cI{\ensuremath{\mathcal{I}}\xspace}
\newcommand\cJ{\ensuremath{\mathcal{J}}\xspace}
\newcommand\cbJ{\ensuremath{\bm{\mathcal{J}}}\xspace}
\newcommand\cK{\ensuremath{\mathcal{K}}\xspace}

\newcommand\cR{\ensuremath{\mathcal{R}}\xspace}
\newcommand\cS{\ensuremath{\mathcal{S}}\xspace}
\newcommand\cT{\ensuremath{\mathcal{T}}\xspace}
\newcommand\cU{\ensuremath{\mathcal{U}}\xspace}
\newcommand\cW{\ensuremath{\mathcal{W}}\xspace}

\usepackage{xspace}

\newcommand{\shi}{$\mathcal{SHI}$\xspace}
\newcommand{\shiq}{$\mathcal{SHIQ}$\xspace}

\newcommand{\sroiq}{$\mathcal{SROIQ}$\xspace}
\newcommand{\sroiqd}{$\mathcal{SROIQ}$(\textbf{D})\xspace}

\newcommand{\trillp}{TRILL$^\text{P}$\xspace}

\newcommand{\bundlen}{BUNDLE\xspace}

\newcommand{\imp}[1]{\textit{#1}}
\newcommand{\vimp}[1]{\textbf{#1}}
\newcommand{\code}[1]{\texttt{#1}}
\newcommand{\prolog}[1]{%
\ifx\@currenvir\@math%
{\normalfont\mathtt{#1}}%
\else%
{\normalfont\texttt{#1}}%
\fi%
}

\newcommand{\owl}[1]{%
\ifx\@currenvir\@math%
{\normalfont\mathtt{#1}}%
\else%
{\normalfont\texttt{#1}}%
\fi%
}

\newcommand{\mc}[1]{\ensuremath{\mathcal{#1}}}

\newcommand{\dlf}[1]{%
\ifx\@currenvir\@math%
{\normalfont\mathsf{#1}}%
\else%
{\normalfont\textsf{#1}}%
\fi%
}%

\newcommand{\new}[1]{#1}
\newcommand{\crash}{{\color{red}crash}\xspace}

\usepackage{array}
\usepackage{multirow} 
\newcolumntype{C}{>{\centering\arraybackslash}m{10.2ex}}
\newcolumntype{Y}{>{\centering\arraybackslash}m{7.5ex}}

\title{A Framework for Reasoning on Probabilistic Description Logics}
\author{Giuseppe Cota\inst{1}\orcidID{0000-0002-3780-6265} \and Riccardo Zese\inst{2}\orcidID{0000-0001-8352-6304} \and  Elena Bellodi\inst{2}\orcidID{0000-0002-3717-3779} \and Evelina Lamma\inst{2}\orcidID{0000-0003-2747-4292} \and Fabrizio Riguzzi\inst{3}\orcidID{0000-0003-1654-9703}}
 \institute{
Dipartimento di Scienze Matematiche, Fisiche e Informatiche -- University of Parma\\
Parco Area delle Scienze, 53/A, 43124 Parma \\\email{giuseppe.cota@unipr.it}\and
Dipartimento di Ingegneria -- University of Ferrara\\
 Via Saragat 1, I-44122, Ferrara, Italy \and
Dipartimento di Matematica e Informatica -- University of Ferrara\\ 
Via Saragat 1, I-44122, Ferrara, Italy
}

\authorrunning{G. Cota, R. Zese et al.}

\begin{document}
\maketitle

\begin{abstract}
While there exist several reasoners for Description Logics, very few of them can cope with uncertainty.
BUNDLE is an inference framework that can exploit several OWL (non-probabilistic) reasoners to perform inference over Probabilistic Description Logics. 

In this chapter, we report the latest advances implemented in BUNDLE. In particular, BUNDLE can now interface with the reasoners of the TRILL system, thus providing a uniform method to execute probabilistic queries using different settings.
BUNDLE can be easily extended and can be used either as a standalone desktop application or as a library in OWL API-based applications that need to reason over Probabilistic Description Logics. 

The reasoning performance heavily depends on the reasoner and method used to compute the probability. We provide a comparison of the different reasoning settings on several datasets.


%

\end{abstract}

\section{Introduction}
\label{sec:intro}

The aim of the Semantic Web is to make information available in a form that is understandable and automatically manageable by machines.
In order to realize this vision, the W3C has supported the development of a family of 
knowledge representation formalisms of increasing complexity for defining ontologies, called OWL (Web Ontology Languages), that are based on Description Logics (DLs). 
Many inference systems, generally called reasoners, have been proposed to reason upon these  ontologies, such as Pellet~\cite{DBLP:journals/ws/SirinPGKK07}, Hermit~\cite{shearer2008hermit} and  Fact++~\cite{tsarkov2006fact++}.

Nonetheless, modeling real-world domains requires dealing with information that is incomplete or that comes from 
sources with different trust levels. This motivates the need for the management of uncertainty in the Semantic Web, and many proposals have appeared for combining probability theory with OWL languages, or with the underlying DLs \cite{DBLP:conf/kr/LutzS10,DBLP:conf/kr/Jaeger94,DBLP:conf/aaai/KollerLP97,Ding04aprobabilistic,DBLP:journals/ai/Lukasiewicz08}. 
Among them, in \cite{RigBelLamZes15-SW-IJ,Zese17-SSW-BK} we introduced the DISPONTE  semantics, which applies the distribution semantics \cite{DBLP:conf/iclp/Sato95} to DLs. 
Examples of systems that perform probabilistic logic inference under DISPONTE are BUNDLE~\cite{RigBelLamZes15-SW-IJ,RigBel15-IJCAI-IC,Zese17-SSW-BK}, and TRILL~\cite{ZesBelRig16-AMAI-IJ,Zese17-SSW-BK,ZesBelCot18-TPLP-IJ}. The former is implemented in Java, the latter in Prolog.


To perform exact probabilistic inference over DISPONTE knowledge bases (KBs), it is necessary to perform either one of the following reasoning tasks: \emph{justification finding} or \emph{pinpointing formula extraction}. The former method consists of finding the covering set of justifications, i.e., the set of all justifications for the entailment of the query. In the latter, a monotonic boolean formula is built, which compactly represents the covering set of justifications.
Both of these non-standard reasoning tasks can be performed by a non-probabilistic OWL reasoner.
The first version of BUNDLE was able to execute justification finding by exploiting the Pellet reasoner only~\cite{DBLP:journals/ws/SirinPGKK07}. 
Then it was extended to exploit different non-probabilistic OWL reasoners and approaches for justification finding~\cite{cotamodularbundle}. In particular, it embeds Pellet, Hermit, Fact++ and JFact as OWL reasoners, and three justification generators, namely GlassBox (only for Pellet), BlackBox and OWL Explanation.

In this chapter, we illustrate the state of the art of the BUNDLE framework. In particular, we present a newer version of BUNDLE which also interfaces with the probabilistic reasoners of the TRILL system, namely: 
\begin{enumerate*}[label=(\roman*)]
	\item TRILL~\cite{ZesBelRig16-AMAI-IJ,Zese17-SSW-BK}, which solves the justification finding problem,
	\item \trillp~\cite{ZesBelRig16-AMAI-IJ,Zese17-SSW-BK}, which returns the pinpointing formula using the approach defined in~\cite{DBLP:journals/jar/BaaderP10,DBLP:journals/logcom/BaaderP10}, and
	\item TORNADO~\cite{ZesBelCot18-TPLP-IJ}, which, similarly to \trillp, returns the pinpointing formula, but the formula is represented in a way that can be directly used to compute the probability. 
\end{enumerate*}
In this way, the user can run probabilistic queries in a uniform way by using the preferred reasoner.
%
%
In addition, BUNDLE can be easily extended by ``plugging-in'' a new reasoner or by including new concrete implementations of algorithms for justification finding/axiom pinpointing.

The performance of reasoning heavily depends on the reasoner and method used to compute the probability for a given query \new{and it is of foremost importance for (distributed) probabilistic rule learning systems such as~\cite{RigBelZes16-ECAI-ICnoser,CotZesBel15-ECMLDC-IW}}. To evaluate the system, we performed several experiments on various real-world and synthetic datasets using different settings.


The chapter is organized as follows: Section~\ref{sec:description-logics} briefly introduces DLs, while Section~\ref{sec:justification_finding} illustrates the problems of justification finding and pinpointing formula extraction. Section~\ref{sec:prob-descr-logics} and Section~\ref{sec:inference} present DISPONTE and the theoretical aspects of inference in DISPONTE KBs respectively. The description of \bundlen is provided in Section~\ref{sec:bundle}. Finally, Section~\ref{sec:exp} shows the experimental evaluation and Section~\ref{sec:conc} concludes the paper.

\section{Description Logics}
\label{sec:description-logics}
An ontology describes the concepts of the domain of interest and their relations with a formalism that allows information to be processable by machines. \new{The \imp{Web Ontology Language} (OWL) is a knowledge representation language for authoring ontologies or knowledge bases. 
The latest version of this language, OWL~2~\cite{owl2:2012}, is a W3C recommendation since 2012. In order to reach some computational properties, it is possible to define a sublanguage of OWL 2, also called \emph{profile}, by applying syntactic restrictions.}

Descriptions Logics (DLs) provide a logical formalism for knowledge representation. They are useful in all the domains where it is necessary to represent information and perform inference on it, such as software engineering, medical diagnosis, digital libraries, databases and Web-based informative systems. They possess nice computational properties such as decidability and (for some DLs) low complexity~\cite{Badeer:2008:DL:52211}.

There are many different DL languages that differ in the constructs that are allowed for defining concepts (sets of individuals of the domain) and roles (sets of pairs of individuals).
The \sroiqd DL is one of the most common fragments; it was introduced by Horrocks et al. in~\cite{horrocks2006even} and it is of particular importance because it is semantically equivalent to OWL 2.

Let us consider a set of \emph{atomic concepts} $\mathbf{C}$, a set of \emph{atomic roles} $\mathbf{R}$ and a set of individuals $\mathbf{I}$.
A \emph{role} could be an atomic role $R \in \mathbf{R}$, the inverse $R^{-}$ of an atomic role $R \in \mathbf{R}$ or a complex role $R \circ S$. We use $\mathbf{R^{-}}$ to denote the set of all inverses of roles in $\mathbf{R}$. 
Each $A \in \mathbf{A}$, $\bot$ and $\top$ are concepts and if $a \in \mathbf{I}$, then $\{a\}$ is a concept called \emph{nominal}. 
If $C$, $C_1$ and $C_2$ are concepts and $R \in \mathbf{R}\cup\mathbf{R^{-}}$, then $(C_1\sqcap C_2)$, $(C_1\sqcup C_2 )$ and $\neg C$ are concepts, as
well as $\exists R.C$, $\forall R.C$, $\exists R.\mathrm{Self}$, $\geq n R.C$ and $\leq n R.C$ for an integer $n \geq 0$. 

A \emph{knowledge base} (KB) $\cK = (\cT, \cR, \cA)$ consists of a TBox $\cT$, an RBox $\cR$ and an ABox $\cA$. An RBox $\cR$ is a finite set of \emph{transitivity axioms} $Trans(R)$, \imp{role asimmetricity axioms} $Asy(R)$, \imp{role disjointness axioms} $Dis(R, S)$, \emph{role inclusion axioms} $R \sqsubseteq S$ and \emph{role chain axioms} $R\circ P \sqsubseteq S$, where $R, P, S \in \mathbf{R} \cup \mathbf{R^{-}}$.
A \emph{TBox} $ \cT $ is a finite set of \textit{concept inclusion axioms} $C\sqsubseteq D$, where $C$ and $D$ are concepts.
An \emph{ABox} $\cA$ is a finite set of \textit{concept membership axioms} $a : C$ and \textit{role membership axioms} $(a, b) : R$, where $C$ is a concept, $R \in \mathbf{R}$ and $a,b \in \mathbf{I}$. 

A KB is usually assigned a semantics using interpretations of the form $\cI = (\Delta^\cI , \cdot^\cI )$, where $\Delta^\cI$ is a
non-empty \emph{domain} and $\cdot^\cI$ is the \emph{interpretation function} that assigns an element in $\Delta ^\cI$ to each individual $a$, a subset of $\Delta^\cI$ to each concept $C$ and a subset of $\Delta^\cI \times \Delta^\cI$ to each role $R$.
The mapping $\cdot^\cI$ is extended to complex concepts as follows (where 
$R^\cI(x,C) = \{y|\langle x,y\rangle\in R^\cI, y \in C^\cI\}$ and $\#X$ denotes the cardinality of the set $X$):
\begin{footnotesize}
$$
\begin{array}{rclcrcl}
\top^\cI&=&\Delta^\cI && \bot^\cI&=&\emptyset\\
\{a\}^\cI&=&\{a^\cI\} && (\neg C)^\cI&=&\Delta^\cI\setminus C^\cI\\
(C_1\sqcup C_2)^\cI&=&C_1^\cI \cup C_2^\cI && (C_1\sqcap C_2)^\cI&=&C_1^\cI \cap C_2^\cI\\
(\exists R.C)^\cI&=&\{x\in \Delta^\cI| R^\cI(x)\cap C^\cI\neq \emptyset\} && (\forall R.C)^\cI&=&\{x\in \Delta^\cI| R^\cI(x)\subseteq C^\cI\}\\
(\geq n R.C)^\cI&=&\{x\in \Delta^\cI| \#R^\cI(x,C)\geq n\} &&
(\leq n R.C)^\cI&=&\{x\in \Delta^\cI| \#R^\cI(x,C)\leq n\}\\
(R^-)^\cI&=&\{\langle y,x \rangle |\langle x,y\rangle \in R^\cI\} &&
(R_1 \circ ... \circ R_n)^\cI &=& R_1^\cI \circ ... \circ R_n^\cI\\
(\exists R.\mathrm{Self})^\cI &=& \{x|\langle x,x \rangle \in R^\cI\}
\end{array}
$$
\end{footnotesize}
\sroiqd also permits the definition of datatype roles, which connect an individual to an element of a datatype such as integers, floats, etc. 

A query $Q$ over a KB $\cK$ is usually an axiom for which we want to test the entailment from the KB, written as $\cK \models Q$. 

\begin{example}
\label{ex:dl_crime}
Consider the following KB ``Crime and Punishment'' containing 4 axioms $\alpha_i$
\begin{align*}
 & \alpha_1 = \dlf{Nihilist} \sqsubseteq \dlf{GreatMan} && \alpha_2 = \exists\dlf{killed}.\top \sqsubseteq \dlf{Nihilist}\\
 & \alpha_3 = \dlf{(raskolnikov, alyona) : killed} && \alpha_4 = \dlf{(raskolnikov, lizaveta) : killed}
\end{align*}
This KB states that if you killed someone then you are a nihilist and whoever is a nihilist is a ``great man'' (TBox). It also states that Raskolnikov killed Alyona and Lizaveta (ABox). The KB entails the query $Q=\dlf{raskolnikov} : \dlf{GreatMan}$ (but are we sure about that?).
\end{example}

\new{
\subsection{Decidability of \sroiq}
In order to ensure decidability of inferencing in \sroiq DL, three conditions must be met:
\begin{itemize}
	\item no cardinality restrictions must be applied on transitive roles or on roles that have transitive subroles~\cite{horrocks1999practical,horrocks1999description};
	\item the RBox must be regular~\cite{horrocks2006even}; 
	\item in the class expressions $\geq n R.C$, $\leq n R.C$ and $\exists R.\mathrm{Self}$, and in the axioms $\exists R.\mathrm{Self} \sqsubseteq \bot$ (also known as \imp{role irreflexivity axiom} $\mathit{Irr}(R)$), $\mathit{Asy}(R)$ and $\mathit{Dis}(R)$, $R$ must be \imp{simple}.
\end{itemize}

\begin{definition}[Regular RBox \cite{horrocks2006even}]
	\begin{itemize}
		\item An RBox is \imp{regular} if there is a strict linear order $\prec$ on roles and the RBox contains only role chain axioms of the following forms:
		\begin{align*}
		& R\circ R \sqsubseteq R && S^- \sqsubseteq R\\
		& S_1 \circ S_2 \circ ... \circ S_n \sqsubseteq R && R \circ S_1 \circ S_2 \circ ... \circ S_n \sqsubseteq R\\
		& S_1 \circ S_2 \circ ... \circ S_n \circ R \sqsubseteq R
		\end{align*}
		where $S_i \prec R$ for all $i= 1, 2,\dots, n$.
	\end{itemize}
\end{definition}

\begin{definition}[Simple role in \sroiqd \cite{horrocks2006even}]
	Given a role $R$, its \imp{simplicity} is inductively defined as follows:
	\begin{itemize}
		\item $R$ is simple if it does not occur on the right hand side of a role inclusion axiom in \mc{R}, i.e. there is no role chain axiom of the form: $S_1 \circ S_2 \circ \dots \circ S_n \sqsubseteq R$;
		\item an inverse role $R^-$ is simple if $R$ is, and
		\item if $R$ occurs on the right hand side of a role inclusion axiom in \mc{R}, then $R$ is simple if, for each $S \sqsubseteq R$, $S$ is a simple role.
	\end{itemize}
\end{definition}
}

\section{Justification Finding and Pinpointing Formula}
\label{sec:justification_finding}

\subsection{Justification Finding}
Here we discuss the problem of finding the covering set of justifications for a given query. This non-standard reasoning service is also known as \imp{axiom pinpointing}~\cite{DBLP:conf/ijcai/SchlobachC03} and it is useful for tracing derivations and debugging ontologies. This problem has been investigated by various authors \cite{Kalyanpurphd,DBLP:conf/ijcai/SchlobachC03,baader2007pinpointing,Horridge09theowl}.
A justification corresponds to an \imp{explanation} for a query $Q$. An explanation is a subset of logical axioms $\cE$ of a KB $\cK$ such that $\cE \models Q$, whereas a justification is an explanation such that it is minimal w.r.t. set inclusion. Formally, we say that an explanation $\cJ\subseteq \cK$ is a justification if for all $\mc{J'} \subset \cJ$, $\mc{J'} \not\models Q$, i.e. $\cJ'$ is not an explanation for $Q$ . The problem of enumerating all justifications that entail a given query is called axiom pinpointing or justification finding. \imp{The set of all the justifications for the query $Q$ is the covering set of justifications for $Q$}. Given a KB $\cK$, the covering set of justifications for $Q$ is denoted by \textsc{All-Just($Q,\cK$)}. 

Below, we provide the formal definitions of justification finding problem. 

%

\begin{definition}[Justification finding problem]
\label{def:minaenum}
\\\textbf{Input}: A knowledge base $\cK$, and an axiom $Q$ such that $\cK \models Q$.\\
\textbf{Output}: The set \textsc{All-Just($Q,\cK$)} of all the justifications for $Q$ in $\cK$.
\end{definition}

\begin{example}
	Consider the KB and the query $Q=\dlf{raskolnikov} : \dlf{GreatMan}$ of Example~\ref{ex:dl_crime}. 
	$\textsc{All-Just($Q,\cK$)} = \{\ \{\alpha_1, \alpha_2, \alpha_3\},\ \{\alpha_1, \alpha_2, \alpha_4\}\ \}.$
	
\end{example}

There are two categories of algorithms for finding a single justification: glass-box~\cite{Kalyanpurphd} and black-box algorithms. The former category is reasoner-dependent, i.e. a glass-box algorithm implementation depends on a specific reasoner, whereas a black-box algorithm is reasoner-independent, i.e. it can be used with any reasoner. In both cases, we still need a reasoner to obtain a justification.

It is possible to incrementally compute all justifications for an entailment by using Reiter's Hitting Set Tree (HST) algorithm~\cite{DBLP:journals/ai/Reiter87}. This algorithm repeatedly calls a glass-box or a black-box algorithm which builds a new justification.
To avoid the extraction of already found justifications, at each iteration the extraction process is performed on a KB from which some axioms are removed by taking into account the previously found justifications. For instance, given a KB \cK and a query $Q$, if the justification $\cJ = \{ \beta_1, \beta_2, \beta_3 \}$ was found, where $\beta_i$s are axioms, to avoid the generation of the same justification, the HST algorithm tries to find a new justification on $\cK' = \cK \smallsetminus \beta_1$. If no new justification is found the HST algorithm backtracks and tries to find another justification by removing other axioms from $\cJ$, one at a time.

\subsection{Pinpointing Formula}
Given a query, instead of finding the covering set of justifications, we can compute the \emph{pinpointing formula}, which compactly represents the covering set of justifications, following the approaches proposed in \cite{DBLP:journals/jar/BaaderP10,DBLP:journals/logcom/BaaderP10}.

To build a pinpointing formula, first we have to associate a unique propositional variable to every axiom $E$ of the KB $\cK$, indicated with $var(E)$. Let $var(\cK)$ be the set of all the propositional variables associated with axioms in $\cK$, then the pinpointing formula is a \emph{monotone Boolean formula} built using some or all of the variables in $var(\cK)$ and the conjunction and disjunction connectives. 
A valuation $\nu$ of a set of variables $var(\cK)$ is the set of propositional variables that are true, i.e., $\nu \subseteq var(\cK)$. For a valuation $\nu \subseteq var(\cK)$, let $\cK_{\nu} := \{E \in \cK |var(E)\in\nu\}$.  

\begin{definition}[Pinpointing formula]
	Given a query $Q$ and a KB $\cK$, a monotone Boolean formula $\phi$ over $var(\cK)$ is called a \emph{pinpointing formula} for $Q$ if for every valuation $\nu \subseteq var(\cK)$ it holds that $\cK_{\nu} \models Q$ iff $\nu$ satisfies $\phi$.
\end{definition}

It is worth noting that a pinpointing formula could be directly generated from the covering set of justifications. However, this formula may not be the most compact.

\begin{example}
	Consider the KB and the query $Q=\dlf{raskolnikov} : \dlf{GreatMan}$ of Example~\ref{ex:dl_crime}. If we associate a Boolean variable $X_i$ with each axiom $\alpha_i$, a pinpointing formula could be
	$ X_1 \wedge X_2 \wedge (X_3 \vee X_4) $.
	Another (non-compact) pinpointing formula could be directly generated from the covering set of justifications: $ (X_1 \wedge X_2 \wedge X_3) \vee (X_1 \wedge X_2 \wedge X_4)$. 
\end{example}

\section{Probabilistic Description Logics}
\label{sec:prob-descr-logics}
DISPONTE \cite{RigBelLamZes15-SW-IJ,Zese17-SSW-BK} applies the distribution semantics \cite{DBLP:conf/iclp/Sato95} to Probabilistic Description Logic KBs. 

In DISPONTE, a \imp{probabilistic knowledge base} \mc{K} is a set of certain axioms and probabilistic axioms. Certain \imp{axioms} take the form of regular DL axioms, whereas \imp{probabilistic axioms} take the form 
$ p :: E $,
where $p \in [0,1]$ and $E$ is a DL axiom. \new{The probability $p$ can be interpreted as an \emph{epistemic probability}, i.e., as the degree of our belief in the truth of axiom $E$. Another interpretation is that $p$ represents the trustworthiness level of the data source for the axiom $E$.}

DISPONTE associates independent Boolean random variables to the DL axioms. The set of axioms that have the random variable assigned to 1 constitutes a \imp{world}. The probability of a world $w$ is computed by multiplying the probability $p_i$ for each probabilistic axiom $\beta_i$ included in the world by the probability $1-p_i$ for each probabilistic axiom $\beta_i$ not included in the world.

Below, we provide some formal definitions for DISPONTE.
%
%

\begin{definition}[Atomic choice]
\label{def:atomic_choice_pdl}
An \imp{atomic choice} is a couple $(\beta_i, k)$ where $\beta_i$ is the $i$th probabilistic axiom and $k \in \{0, 1\}$. The variable $k$ indicates whether $\beta_i$ is chosen to be included in a world $(k = 1)$ or not $(k = 0)$.
\end{definition}

\begin{definition}[Composite choice]
A \imp{composite choice} $\kappa$ is a consistent set of atomic choices, i.e., $(\beta_i, k) \in \kappa$, $(\beta_i, m) \in \kappa$ implies
$k = m$ (only one decision is taken for each axiom).
\end{definition}
The probability of composite choice $\kappa$ is: 
$P(\kappa) = \prod_{(\beta_i, 1)\in \kappa}p_i\prod_{(\beta_i, 0)\in \kappa}(1-p_i)$,
where $p_i$ is the probability associated with axiom $\beta_i$, because the random variables associated with axioms are independent.

\begin{definition}[Selection]
A \imp{selection} $\sigma$ is a total composite choice, i.e., it contains an atomic choice $(\beta_i, k)$ for every probabilistic axiom of the theory. A \imp{selection} $\sigma$ identifies a theory $w_\sigma$ called a \imp{world}: $w_\sigma = \mathcal{C}\cup\{\beta_i|(\beta_i, 1)\in \sigma\}$, where \mc{C} is the set of certain axioms. 
\end{definition}


$P(w_\sigma)$ is a probability distribution over worlds. Let us indicate with \cW the set of all worlds. The probability of Q is~\cite{RigBelLamZes15-SW-IJ}:
$
P(Q) = \sum_{w\in \cW : w\models Q}P(w)
$,
i.e. the probability of the query is the sum of the probabilities of the worlds in which the query is true.


\begin{example}
\label{ex:pdl_crime}
 Let us consider the knowledge base and the query $Q=\dlf{raskolnikov} : \dlf{GreatMan}$ of Example~\ref{ex:dl_crime} where some of the axioms are probabilistic:
 \begin{align*}
 \beta_1 =&\ 0.2 :: \dlf{Nihilist} \sqsubseteq \dlf{GreatMan} & \alpha_1=&\ \exists\dlf{killed}.\top \sqsubseteq \dlf{Nihilist}\\
 \beta_2 =&\ 0.6 :: \dlf{(raskolnikov, alyona) : killed} & \beta_3=&\ 0.7 :: \dlf{(raskolnikov, lizaveta) : killed}
\end{align*}
Whoever is a nihilist is a ``great man'' with probability 0.2 ($\beta_1$) and Raskolnikov killed Alyona and Lizaveta with probability 0.6 and 0.7 respectively ($\beta_2$ and $\beta_3$). Moreover there is a certain axiom ($\alpha_1$).
The KB has eight worlds and $Q$ is true in three of them, corresponding to the selections: 
\begin{align*}
\{\ \{(\beta_1,1),(\beta_2,1),(\beta_3,1)\},\ \{(\beta_1,1),(\beta_2,1),(\beta_3,0)\},\ \{(\beta_1,1),(\beta_2,0),(\beta_3,1)\}\ \}
\end{align*}
The probability is 
$P(Q) = 0.2\cdot 0.6\cdot 0.7 + 0.2\cdot0.6\cdot(1-0.7) + 0.2\cdot (1-0.6)\cdot0.7 = 0.176$.
\end{example}

\section{Inference in Probabilistic Description Logics}
\label{sec:inference}
It is often infeasible to find all the worlds where the query is true. To reduce reasoning time, inference algorithms find, instead, explanations for the query and then compute the probability of the query from them. Below we provide the definitions of DISPONTE explanations and justifications, which are tightly intertwined with the previous definitions of explanation and justification for the non-probabilistic case.
\begin{definition}[DISPONTE Explanation]
\label{def:expl_pdls}
A composite choice $\phi$ \imp{identifies} a set of worlds $\omega_\phi = \{ w_\sigma | \sigma \in \cS, \sigma\supseteq \phi\}$, where \cS is the set of all selections. We say that $\phi$ is an explanation for $Q$ if $Q$ is entailed by every world of $\omega_\phi$. 
\end{definition}
\begin{definition}[DISPONTE Justification]
We say that an explanation $\gamma$ is a justification if, for all $\gamma' \subset \gamma$, $\gamma'$ is not an explanation for $Q$.
\end{definition}
%

The set of worlds $\omega_{\bm{\Phi}}$ identified by the set of explanations $\bm{\Phi}$ is \imp{covering} $Q$ if every world $w_\sigma \in \cW$ in which $Q$ is entailed is such that 
$w_\sigma \in \omega_{\bm{\Phi}}$.
In other words, a covering set $\bm{\Phi}$ identifies all the worlds in which $Q$ succeeds.

Two composite choices $\kappa_1$ and $\kappa_2$ are \imp{incompatible} if their union is inconsistent. For example, $\kappa_1 = \{(\beta_i, 1) \}$ and $\kappa_2 = \{(\beta_i, 0)\}$ are incompatible. A set $\bm{K}$ of composite choices is \imp{pairwise incompatible} if for all $ \kappa_1 \in \bm{K}$, $\kappa_2 \in \bm{K}$, $\kappa_1 \ne \kappa_2$ implies that $\kappa_1$ and $\kappa_2$ are incompatible. 
The \imp{probability of a pairwise incompatible set of composite choices} $K$ is $P(\bm{K}) = \sum_{\kappa \in \bm{K}} P(\kappa)$.

Given a query $Q$ and a covering set of pairwise incompatible explanations $\bm{\Phi}$, the probability of $Q$ is~\cite{RigBelLamZes15-SW-IJ}:
\begin{equation}
\label{eq:covering_pairwise_incompatible}
P(Q) = \sum_{w_\sigma \in \omega_{\bm{\Phi}}} P(w_\sigma) = P(\omega_{\bm{\Phi}}) = P(\bm{\Phi}) = \sum_{\phi \in \bm{\Phi}} P(\phi)
\end{equation}


Unfortunately, in general, explanations (and hence justifications) are not pairwise incompatible, therefore the covering set of justifications cannot be directly used to compute the probability. Even more so, the pinpointing formula, which compactly encodes the covering set of justifications, cannot be directly used for probability computation. The problem of calculating the probability of a query is therefore reduced to that of finding a covering set of justifications or the pinpointing formula and then transforming it into a covering set of pairwise incompatible explanations. 

We can think of using non-probabilistic reasoners for justification finding or pinpointing formula extraction, then consider only the probabilistic axioms and transform the covering set of justifications or the pinpointing formula into a pairwise incompatible covering set of explanations from which it is easy to compute the probability.



\begin{example}
Consider the KB and the query $Q=\dlf{raskolnikov} : \dlf{GreatMan}$ of Example~\ref{ex:pdl_crime}. If we use justification finding algorithms by ignoring the probabilistic annotations, we find the following non-probabilistic justifications:
$ \cbJ = \{\ \{\beta_1, \alpha_1, \beta_2\},\ \{\beta_1, \alpha_1, \beta_3\}\ \} $.
Then we can translate them into DISPONTE justifications:
$\bm{\Gamma} = \{\ \{(\beta_1, 1), (\beta_2, 1)\},\ \{(\beta_1, 1), (\beta_3, 1)\}\ \}$.
 Note that $\bm{\Gamma}$ is not pairwise incompatible, therefore we cannot directly use Equation~(\ref{eq:covering_pairwise_incompatible}). The solution to this problem will be shown in the following section.
\end{example}

\section{BUNDLE}
\label{sec:bundle}
The reasoner \bundlen~\cite{RigBelLamZes15-SW-IJ,RigBel15-IJCAI-IC} computes the probability of a query w.r.t. DISPONTE KBs by first computing all the justifications for the query, then converting them into a pairwise incompatible covering set of explanations by building a Binary Decision Diagram (BDD). Finally, it computes the probability by traversing the BDD using function \textsc{Probability} described in \cite{DBLP:journals/tplp/KimmigDRCR11}. A BDD for a function of Boolean variables is a rooted graph that has one level for each Boolean variable. A node $n$  has two children corresponding respectively to the 1 value and the 0 value of the variable associated with the level of $n$. When drawing BDDs, the 0\--branch is distinguished from the 1-branch by drawing it with a dashed line. The leaves store either 0 or 1.

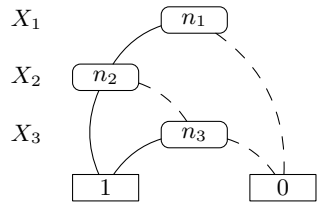
\begin{figure}[ht]
	$$
 \xymatrix@=3mm
 { 
 X_{1} & &*=<25pt,10pt>[F-:<3pt>]{n_1}
 \ar@/_/@{-}[ld] \ar@/^1pc/@{--}[dddr]\\ 
 X_{2}  & *=<25pt,10pt>[F-:<3pt>]{n_2} 
 \ar@/_/@{-}[dd]\ar@/^/@{--}[dr] 
 \\
 X_{3} & & *=<25pt,10pt>[F-:<3pt>]{n_3}
 \ar@/_/@{-}[dl] \ar@/^/@{--}[dr]  
 \\
 &*=<25pt,10pt>[F]{1} &&*=<25pt,10pt>[F]{0}
 }
 $$
	
 \caption{BDD representing the set of explanations for the query of Example \ref{ex:pdl_crime}.}
 \label{dd}
 \end{figure}

\begin{example}[Example~\ref{ex:pdl_crime} cont.] 
\label{ex:pdl_crime_bdd}
 Let us consider the KB and the query of Example~\ref{ex:pdl_crime}.
If we associate random variables $X_{1}$ with axiom $\beta_1$, $X_{2}$ with $\beta_2$ and $X_3$ with $\beta_3$, the BDD representing the set of explanations is shown in Figure \ref{dd}.
By applying function \textsc{Probability} \cite{DBLP:journals/tplp/KimmigDRCR11} to this BDD we get
\begin{eqnarray*}
\mbox{\textsc{Probability}}(n_3)&=&0.7\cdot 1+0.3\cdot 0=0.7\\
\mbox{\textsc{Probability}}(n_2)&=&0.6\cdot 1+0.4\cdot 0.7=0.88\\
\mbox{\textsc{Probability}}(n_1)&=&0.2\cdot 0.88+0.8\cdot 0=0.176
\end{eqnarray*}
and therefore $P(Q)=\mbox{\textsc{Probability}}(n_1)=0.176$, which corresponds to the probability given by DISPONTE.
\end{example}

BUNDLE uses implementations of the HST algorithm to incrementally obtain all the justifications from non-probabilistic reasoner. Moreover, in the latest version, BUNDLE can exploit the reasoners provided by the TRILL framework, thus providing a simple and uniform way to execute probabilistic queries by using the preferred method and reasoner. 

\begin{figure}
 \centering
 \includegraphics[width=0.83\textwidth]{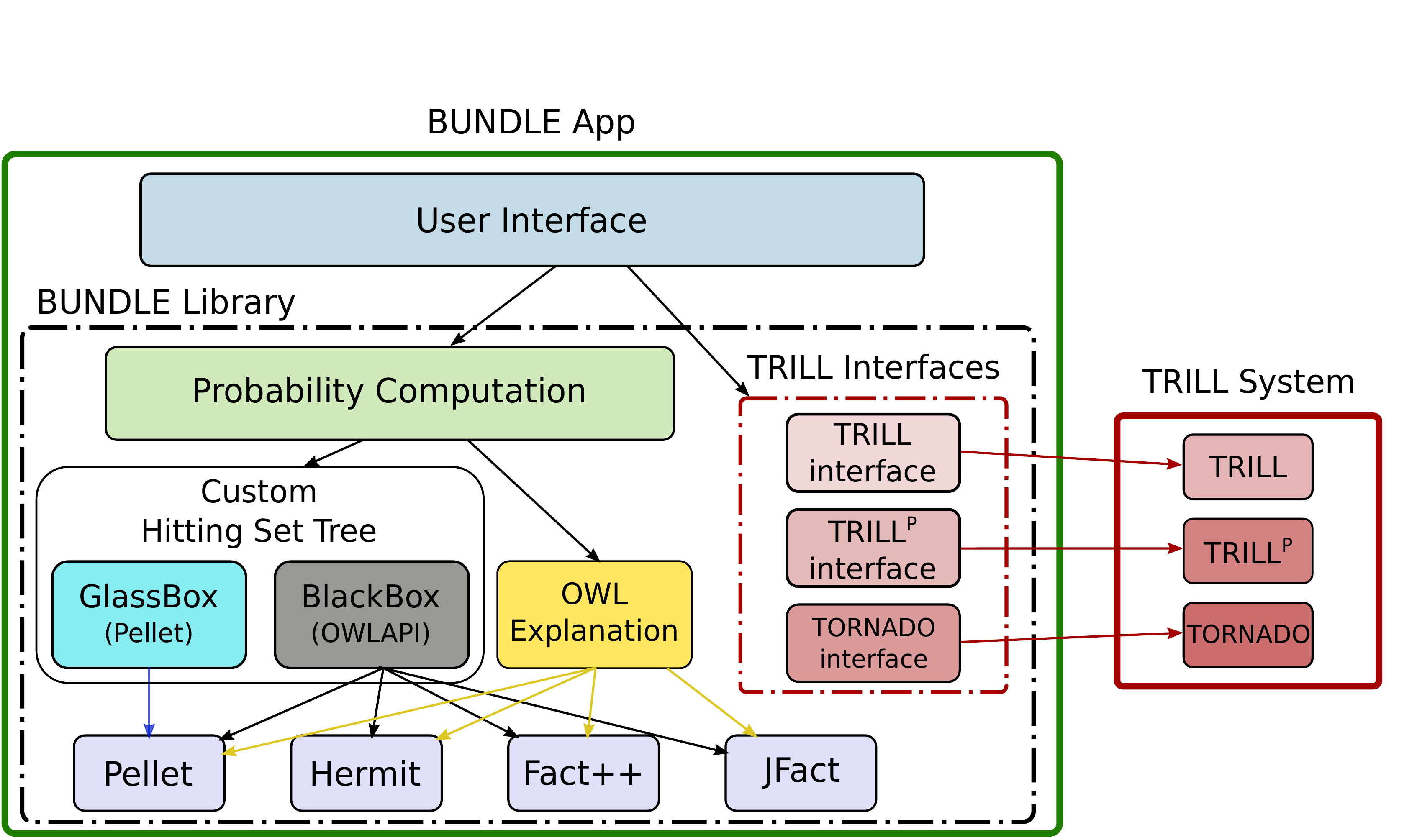}
\caption{Software architecture of \bundlen.}
\label{fig:bundle_arch}
\end{figure}

Figure~\ref{fig:bundle_arch} shows the new architecture of \bundlen. 
The main novelties are the interfaces for the probabilistic reasoners of the TRILL system. These reasoners directly return the probability of the query, therefore, when using one of the TRILL reasoner interfaces, the Probability Computation module, which converts a covering set of justifications into a BDD, is not used.

BUNDLE now supports: 
\begin{enumerate*}[label=(\arabic*)]
	\item four different OWL reasoners: Pellet 2.5.0, Hermit 1.3.8.413~\cite{shearer2008hermit}, Fact++ 1.6.5~\cite{tsarkov2006fact++}, and JFact 4.0.4\footnote{\url{http://jfact.sourceforge.net/}};
	\item three probabilistic reasoners TRILL, \trillp and TORNADO, which exploit Prolog's backtracking feature to obtain the justifications or the pinpointing formula; and
	\item three  different strategies for finding a justification using a non probabilistic reasoner, which are:
\end{enumerate*}
\begin{description}
 \item[GlassBox] A glass-box approach which depends on Pellet. It is a modified version of the \code{GlassBoxExplanation} class contained in the Pellet Explanation library.
 \item[BlackBox] A black-box approach offered by the OWL API\footnote{\url{http://owlcs.github.io/owlapi/}}~\cite{horridge2011owl}. The OWL API is a Java API for the creation and manipulation of OWL 2 ontologies.
 \item[OWL Explanation] A library that is part of the OWL Explanation Workbench~\cite{Horridge09theowl}. The latter  also contains a Protégé plugin, underpinned by the library, that allows Protégé users to find justifications for entailments in their OWL 2 ontologies.
\end{description}
All the supported non-probabilistic reasoners can be paired with the BlackBox or the OWL Explanation methods, while only Pellet can exploit the GlassBox method.

To find all justifications using the GlassBox and BlackBox approaches, we extended the \code{HSTExplanationGenerator} class of the OWL API, which provides an implementation of the HST algorithm, in order to support annotated axioms (DISPONTE axioms are OWL axioms annotated with a probability). OWL Explanation, instead, already contains an HST implementation and a black-box approach that supports annotated axioms.

Table~\ref{table:reasoners} provides an overview of the reasoners and methods supported by the BUNDLE framework. 

\begin{table}
	\caption{Reasoners supported by the BUNDLE framework. The symbol \checkmark means that the reasoner is compatible with the method used for computing the probability.}
	\label{table:reasoners}  
	\fontsize{7.5}{9}\selectfont
	\begin{tabular}{l|l|c|c|c|c|c|}
		\cline{3-7}
		\multicolumn{2}{c|}{} & \multicolumn{4}{c|}{\bfseries Justification Finding} & \textbf{Pinpointing Formula}\\
		\cline{3-7}
		\multicolumn{2}{c|}{} & \multicolumn{3}{c|}{\bfseries Hitting Set Tree} & \multicolumn{2}{c|}{\textbf{Prolog backtracking}}\\
		\hline
		\textbf{Reasoner}& \textbf{DL} & \textbf{GlassBox} & \textbf{BlackBox} & \textbf{OWL Expl.} & \textbf{built-in} & \textbf{built-in}\\		
		\hline\hline
		Pellet & \sroiqd & \checkmark & \checkmark & \checkmark & & \\
		Hermit & \sroiqd & & \checkmark & \checkmark & & \\
		Fact++ & \sroiqd & & \checkmark & \checkmark & & \\
		JFact & \sroiqd & & \checkmark & \checkmark & & \\
		TRILL & \shiq & & & & \checkmark & \\
		\trillp  & \shi & & & & & \checkmark\\
		TORNADO &\shi & & & & & \checkmark\\
		\hline 
	\end{tabular}
\end{table}

BUNDLE can be easily extended in three main ways: 
\begin{itemize}
	\item By adding a non-probabilistic reasoner that implements the \verb|OWLReasoner| interface of the OWL API library.
	\item By adding a probabilistic reasoner that implements the interface \linebreak\verb|ProbabilisticReasoner| defined in BUNDLE. Indeed, the interfaces for the TRILL reasoners implement this Java interface.
	\item By adding a non-probabilistic reasoner that is able to perform the reasoning task of justification finding. This reasoner should implement the \verb|ExplanationReasoner| interface defined in BUNDLE.
\end{itemize}

BUNDLE can be used as standalone desktop application or as a library. \new{Moreover, we developed a web application available at \url{http://bundle.ml.unife.it/} in a similar way to what has been done for TRILL~\cite{BelLamRig16-SPE-IJ} and \texttt{cplint}~\cite{AlbCotRigZes16-AIIA-IC}. The web application allows the user to test BUNDLE without installing any software on the local machine.}

\subsection{Using BUNDLE as an Application}
BUNDLE is an open-source software and is available on Bitbucket, together with its manual, at  \url{https://bitbucket.org/machinelearningunife/bundle}. 

A \bundlen image was deployed in Docker Hub. Users can start using BUNDLE by executing the following commands:
\begin{small}
\begin{lstlisting}
 sudo docker pull giuseta/bundle:4.0.0
 sudo docker run -it giuseta/bundle:4.0.0 bash
\end{lstlisting}
\end{small}
A bash shell of the container then starts and users can execute probabilistic queries by running the command \code{bundle}.

\subsection{Using BUNDLE as a Library}
\bundlen can also be used as a library. 
Once the developer has added  \bundlen dependency in the project's POM file, the probability of the query can be obtained in just few lines:
{\fontsize{8}{8}\selectfont
\begin{lstlisting}[language=Java,xleftmargin=0.1\textwidth, numbers=left]
BundleConfigurationBuilder configBuilder = new BundleConfigurationBuilder(ontology);
BundleConfiguration config = configBuilder
  .hstMethod(hstMethod).reasoner(reasonerName)
  .buildConfiguration();
Bundle reasoner = new Bundle(config);
reasoner.init();
QueryResult result = reasoner.computeQuery(query);
\end{lstlisting}
}
where \texttt{ontology} and \texttt{query} are objects of the classes \texttt{OWLOntology} and \texttt{OWLAxiom} of the OWL API library respectively. 

Lines 1-4 show that the developer can inject the preferred HST method for justification finding (none for the TRILL reasoners) and the favorite reasoner by using a configuration builder. After initialization (lines 5-6), probabilistic inference is performed (line 7).

\section{Experiments}
\label{sec:exp}
We performed four different tests to compare the possible configurations of BUNDLE, which depend on the reasoner and the justification search strategy chosen, for a total of 12 combinations. For each query we set a timeout of 10 minutes. In the first test we compared all configurations on four different datasets, in order to highlight which combination reasoner/strategy shows the best behavior in terms of inference time. To investigate the scalability of the different configurations, in the last three experiments, we considered KBs of increasing size in terms of the number of probabilistic axioms. 

All tests were performed on the HPC System Galileo\footnote{\url{http://www.hpc.cineca.it/hardware/galileo}} equipped with Intel Xeon E5-2697 v4 (Broadwell) @ 2.30 GHz, using 1 core for each test.

\paragraph{Test 1}
\label{par:test_1}
The first test considers 4 real world KBs of various complexity as in~\cite{ZesBelRig16-AMAI-IJ}:
(1)~\textbf{BRCA}~\cite{DBLP:conf/semweb/KlinovP08}, which models the risk factors of breast cancer; 
 (2)~an extract of  \textbf{DBPedia}
 \cite{dbpedia-swj}, containing structured information from Wikipedia, usually those contained in the information box on the righthand side of pages; 
 (3)~\textbf{Biopax level 3}
 \cite{demir2010biopax}, which models metabolic pathways; 
 (4)~\textbf{Vicodi}
 \cite{DBLP:journals/lalc/NagypalDO05}, which contains information on European history and models historical events and important personalities. 

\new{We used a version of the DBPedia, Biopax and BRCA KBs without the ABox and a version of Vicodi with an ABox of 19 individuals.
For each KB we added a probability annotation to each axiom. The probabilistic values of a KB were randomly assigned using a uniform distribution $\cU(0,1)$ and are fixed for all the queries performed on that KB. 
We randomly created 50 different subclass-of queries for BRCA, DBPedia and BioPax, and 50 different instance-of queries for Vicodi, following the concepts hierarchy of the KBs, ensuring each query had at least one explanation.}

Table \ref{table:res_1} shows the average time in seconds to answer queries  with different BUNDLE configurations. Bold values highlight the fastest configuration for each KB. Cells with ``--'' indicate that the timeout was reached in at least one query. Cells with ``\crash'', instead, indicate that the reasoner experienced an internal error and was not able to return a result.

Overall, the best results are obtained by Pellet with the GlassBox approach. However, the use of OWL Explanation library shows competitive results. For BioPax and Vicodi KBs, the Fact++/BlackBox configuration was not able to return a result for any query. TRILL and \trillp reached the timeout at least in one query for BioPax and BRCA, while using TORNADO a timeout occurred when querying BioPax.

\begin{table}[ht]
 \caption{Average time (in seconds) for probabilistic inference with all possible configurations of \bundlen over different datasets (\nameref{par:test_1}). ``--'' means that the execution timed out (600 s). \crash means that the reasoner experienced an internal error.}
\label{table:res_1}  
  \centering
	\fontsize{8}{9}\selectfont
\begin{tabular}{ll|c|c|c|c||c}
  & & \multicolumn{4}{c||}{\textbf{Dataset}} & \\
\textbf{Reasoner} & \textbf{Method} & \textbf{BioPax}  & \textbf{BRCA} & \textbf{DBPedia} & \textbf{Vicodi} & \textbf{Average}\\
\hline\hline
Pellet	& GlassBox 	& 1.316			& 0.886			& 0.623			& 0.956 		& \vimp{0.945} \\
Pellet	& BlackBox	& 1.619			& 1.653			& 0.570			& 1.614 		& 1.364 \\
Pellet	& OWLExp 	& 1.070			& 1.034			& 0.914			& 1.420 		& 1.110 \\
Hermit	& BlackBox	& 4.199			& 6.694			& 1.299			& 4.832			& 4.256 \\
Hermit	& OWLExp	& 1.198			& 2.071			& 0.835			& 2.024			& 1.532 \\
JFact	& BlackBox	& 1.648			& 1.852			& 0.541			& 1.341			& 1.346 \\
JFact	& OWLExp	& \vimp{0.887}	& 1.012			& 0.878			& 1.179			& 0.989 \\
Fact++	& BlackBox	& \crash		& 0.649			& 0.363			& \crash		& n.a. \\
Fact++	& OWLExp	& 0.903			& \vimp{0.554}	& 0.588			& 3.285			& 1.333 \\
TRILL	& 			& --			& --			& 0.442			& 0.470			& n.a. \\
\trillp	&			& --			& --			& 0.335			& 0.424			& n.a.	\\
TORNADO	& 			& --			& 2.439			& \vimp{0.287}	& \vimp{0.373}	& n.a.\\
\hline
\end{tabular}
\end{table}

\paragraph{Test 2}\label{par:test_2}
The second test was performed following the approach presented in \cite{DBLP:conf/semweb/KlinovP08} on the BRCA KB ($\mathcal{ALCHF}(D)$, 490 axioms). 
To test \bundlen, we randomly generated and added an increasing number of subclass-of probabilistic axioms. The number of these axioms was varied from 9 to 16, and, for each number, 30 different consistent KBs were created. \new{Every time a KB is generated, the probability values of the axioms are generated anew}. The number of additional axioms may cause an exponential increase of the inference complexity (please see \cite{DBLP:conf/semweb/KlinovP08} for a detailed explanation). \new{In these tests we consider those possible cases where most of our knowledge is certain but there are some uncertainties.}

Finally, an individual was added to every KB, randomly assigned to each simple class that appeared in the probabilistic axioms, and a random  probability was attached to it. We ran 60 probabilistic queries of the form $a : C$ where $a$ is the added individual and $C$ is a class randomly selected among those that represent women under increased and lifetime risk such as \textit{WomanUnderLifetimeBRCRisk} and \textit{WomanUnderStronglyIncreasedBRCRisk}, which are at the top of the concept hierarchy.

Table \ref{tab:increasing_BRCA} shows the execution time averaged over the 60 queries as a function of the number of probabilistic axioms. For each size, bold values indicate the best configuration. The results show that TORNADO is the only reasoner that can provide answers by respecting the time limits and all its execution timings where competitive with the best configurations.


\begin{table}[htb]
 \caption{Average execution time (in seconds) for probabilistic inference with different configurations of \bundlen on versions of the BRCA KB of increasing size (\nameref{par:test_2}). ``--'' means that the execution  timed out (600 s).}
 \label{tab:increasing_BRCA}  
    \centering
	\fontsize{8}{9}\selectfont
    \begin{tabular}{ll|Y|Y|Y|Y|Y|Y|Y|Y}
    \textbf{Reasoner} & \textbf{Method} & \vimp{9} & \vimp{10} & \vimp{11} & \vimp{12} & \vimp{13} & \vimp{14} & \vimp{15} & \vimp{16}\\
    \hline\hline
Pellet	& GlassBox	& \vimp{2.437}	& \vimp{1.812}	& 14.914		& --			& \vimp{4.101}	& --			& \vimp{3.743}	& 7.102			\\
Pellet	& BlackBox	& 8.082			& 5.152			& 29.151		& --			& 14.470		& --			& 12.453		& 18.727		\\
Pellet	& OWLExp	& 3.034			& 2.339			& 10.507		& --			& 4.759			& --			& 4.025			& 5.405			\\
Hermit	& BlackBox	& 38.431		& 22.625		& --			& --			& 73.854		& --			& --			& --			\\
Hermit	& OWLExp	& 11.249		& 8.497			& 29.536		& --			& 18.850		& --			& --			& 19.692		\\
JFact	& BlackBox	& 8.937			& 5.759			& 35.504		& --			& 16.451		& --			& 15.096		& 23.108		\\
JFact	& OWLExp	& 3.068			& 2.384			& 8.494			& --			& 4.493			& --			& 3.978			& 5.103			\\
Fact++	& BlackBox	& --			& --			& 17.015		& --			& --			& --			& --			& --			\\
Fact++	& OWLExp	& --			& --			& --			& --			& --			& --			& --			& \vimp{4.510}	\\
TRLL	&			& --			& --			& --			& --			& --			& --			& --			& --			\\
\trillp	&			& --			& --			& --			& --			& --			& --			& --			& --			\\
TORNADO	&			& 3.425			& 2.659			& \vimp{4.705}	& \vimp{4.252}	& 4.604			& \vimp{4.277}	& 4.022			& 4.745			\\
    \hline
    \end{tabular}

\end{table}

\paragraph{Test 3}\label{par:test_3}
In the third test we artificially created a set of KBs of increasing size of the following form:
\begin{align*}
(\beta_{1,i})\ 0.6 :: B_{i-1}\sqsubseteq P_i\sqcap Q_i & & (\beta_{2,i})\ 0.6 :: P_i\sqsubseteq B_i && (\beta_{3,i})\ 0.6 :: Q_i\sqsubseteq B_i
\end{align*}
where $n \geq 1$ and $1\leq i\leq n$. The query $Q= B_0\sqsubseteq B_n$ has $2^n$ justifications, even if the KB has a size that is linear in $n$.
We increased $n$ from 2 to 10 in steps of 2 and we collected the running time, averaged over 50 executions.
Table \ref{table:res_3} shows, for each $n$, the average time in seconds that the systems took for computing the probability of the query $Q$ (in bold the best time for each size). Cells with ``--'' indicate that the timeout occurred at least in one query.

The experimental results show that using TORNADO outperforms the other settings. If the size of the synthetic dataset is greater or equal to 10, all the runs reach a timeout.

\begin{table}
 \caption{Average execution time (in seconds) for probabilistic inference with different configurations of BUNDLE on synthetic datasets (\nameref{par:test_3}). ``--'' means that the execution  timed out (600 s).}
 \label{table:res_3}  
    \centering
	\fontsize{8}{9}\selectfont
    \begin{tabular}{ll|C|C|C|C|C}
    \textbf{Reasoner} & \textbf{Method} & \vimp{2} & \vimp{4} & \vimp{6} & \vimp{8} & \vimp{10}\\
    \hline\hline
Pellet 	& GlassBox	& 0.890			& 1.573			& 7.720			& --				& --\\
Pellet	& BlackBox	& 1.060			& 2.446			& 12.738		& -- 				& --\\
Pellet	& OWLExp	& 1.843			& 4.055			& 10.443		& 31.118			& --\\
Hermit	& BlackBox	& 5.968			& 29.316		& 168.286		& -- 				& --\\
Hermit	& OWLExp	& 4.410			& 16.637		& 63.062		& 239.256			& --\\
JFact	& BlackBox	& 1.039			& 2.205			& 10.978		& --				& --\\
JFact	& OWLExp	& 1.749			& 3.869			& 9.833			& 28.170			& --\\
Fact++	& BlackBox	& --			& 2.229			& --			& --				& --\\
Fact++	& OWLExp	& 1.728			& --			& 10.159		& 30.923			& --\\
TRILL	&			& 0.549			& 1.244			& 3.708			& 34.105			& --\\
\trillp	&			& 0.267			& 0.443			& 23.767		& --				& --\\
TORNADO	& 			& \vimp{0.194}	& \vimp{0.203}	& \vimp{0.240}	& \vimp{0.343}	& --\\
\hline
    \end{tabular}
\end{table}

\paragraph{Test 4}\label{par:test_4}
To further test the various settings of the BUNDLE framework on real world KBs, we have conducted a test using the \emph{Foundational Model of Anatomy Ontology} (FMA for short)\footnote{\url{http://si.washington.edu/projects/fma}}.
FMA is a KB for biomedical informatics that models the phenotypic structure of the human body anatomy. \new{It contains 4,706 axioms in the TBox and RBox, with 2,626 different classes. To perform this test, we created 7 versions of the KB containing an increasing number of individuals. The first 5 versions contains a number of individuals varying from 20 to 100 with steps of 20, while the last 2 versions contain 200 and 300 individuals. Each individual can be the subject of 1 to 11 assertions. Then we added a probability annotation to each axiom with random values sampled from a uniform distribution $\cU(0,1)$. For each KB of a given size, we ran 10 times the query $i_0 :$ \dlf{Organ\_zone}, where $i_0$ is an individual which is present in all the KBs. The averaged running time is reported in Table~\ref{table:res_4}.}

The results show that Pellet with GlassBox obtains the best performances and that with OWL Explanation paired with any reasoner we obtain competitive results. \new{However, it seems that the Prolog-based approaches have some issues handling high numbers of individuals}.

\begin{table}
	\caption{Average execution time (in seconds) for probabilistic inference with different configurations of BUNDLE on versions of the FMA KB of increasing size (\nameref{par:test_4}). ``--'' means that the execution  timed out (600 s).}
	\label{table:res_4}  
	\centering
	\fontsize{8}{9}\selectfont
	\begin{tabular}{ll|c|c|c|c|c|c|c}
		\textbf{Reasoner} & \textbf{Method} & \vimp{20}	& \vimp{40} & \vimp{60} & \vimp{80} & \vimp{100} & \vimp{200} & \vimp{300}\\
		\hline\hline

Pellet	& GlassBox	& \vimp{1.392}  & \vimp{1.491}  	& \vimp{1.611}	& \vimp{1.594} 	& \vimp{2.457} 	& 1.890  		& 1.931 \\
Pellet 	& BlackBox	& 9.253  		& 9.458 			& 11.772		& 15.509 		& 43.435		& 20.649 		& 22.118 \\
Pellet	& OWLExp 	& 2.000  		& 2.004 			& 2.091			& 2.092			& 5.757 		& 2.371			& 2.470 \\
Hermit	& BlackBox	& 23.660 		& 24.306			& 26.570 		& 32.344	 	& 97.917		& 40.773 		& 40.969 \\
Hermit	& OWLExp 	& 3.913			& 3.907 			& 3.950 		& 4.201		 	& 10.349	 	& 4.169 		& 4.184 \\
JFact	& BlackBox 	& 7.715  		& 8.206			 	& 9.612		 	& 12.129	 	& 31.382		& 17.708 		& 17.547 \\
JFact 	& OWLExp 	& 1.900			& 1.985			 	& 1.956		 	& 2.003 		& 4.995		 	& 2.290 		& 2.294 \\
Fact++	& BlackBox 	& 8.219			& --		 		& 9.526 	 	& 11.676	 	& 34.216	 	& 16.953 		& -- \\
Fact++ 	& OWLExp	& 1.644 		& 1.687				& 1.689		 	& 1.774			& 4.648 	 	& \vimp{1.853}	& \vimp{1.845} \\
TRILL	& 			& 17.286		& 23.912		 	& 35.932 		& 59.228		& 93.456 		& --	 		& -- \\
\trillp	& 			& 17.884		& 31.431			& 50.811	 	& 89.426	 	& 142.585	 	& -- 			& -- \\
TORNADO	&			& 17.035		& 24.192		 	& 37.112 		& 62.409	 	& 97.258	 	& -- 			& -- \\
		\hline
	\end{tabular}
\end{table}

\section{Conclusions}
\label{sec:conc}
In this chapter, we illustrated the state of the art of BUNDLE, a framework for reasoning on Probabilistic Description Logics KBs that follow DISPONTE. The framework can be used both as a standalone application and as a library and it allows to pair 4 different OWL reasoners with 3 different approaches to find query justifications. Moreover, the latest version add the possibility of using the Prolog-based probabilistic reasoners TRILL, \trillp and TORNADO. \new{We also developed a web application that allows to test BUNDLE without installing any software on the local machine.}

We provided a comparison between the various configurations reasoner/approach over different datasets, showing that in 2 out of 4 experiments (\nameref{par:test_1} and \nameref{par:test_4}) Pellet paired with GlassBox or any reasoner paired with the OWLExplanation library achieve the best results in terms of inference time on a probabilistic ontology. However, for BRCA datasets of increasing size (\nameref{par:test_2}) and for synthetic datasets with queries that have an exponential number of justifications (\nameref{par:test_3}), TORNADO showed the best performance. 

In the future, we plan to study the effects of glass-box or grey-box methods for collecting explanations. \new{Moreover, we plan to integrate in BUNDLE a reasoner based on Abductive Logic Programming~\cite{DBLP:conf/iclp/GavanelliLRBZC15}.}

\textbf{Acknowledgement} This work was supported by the ``GNCS-INdAM''.


\bibliographystyle{splncs04}
\bibliography{bibliography/journals_short,bibliography/booktitles_long,bibliography/booktitles_springer,bibliography/series_long,bibliography/series_springer,bibliography/publishers_long,bibliography/bibl}

\end{document}